\documentclass[11pt,a4paper]{article}
\usepackage[hyperref]{acl2021}
\usepackage{times}
\usepackage{latexsym}

\def\scasp{s(CASP) }

\usepackage{ifxetex,ifluatex}
\usepackage{fixltx2e} 
\IfFileExists{upquote.sty}{\usepackage{upquote}}{}
\ifnum 0\ifxetex 1\fi\ifluatex 1\fi=0 
  \usepackage[utf8]{inputenc}
\else 
  \ifxetex
    \usepackage{mathspec}
    \usepackage{xltxtra,xunicode}
  \else
    \usepackage{fontspec}
  \fi
  \defaultfontfeatures{Mapping=tex-text,Scale=MatchLowercase}
  
\fi
\IfFileExists{microtype.sty}{\usepackage{microtype}}{}
\usepackage{color}
\usepackage{fancyvrb}

\DefineVerbatimEnvironment{HighlightingFancy}{Verbatim}{commandchars=\\\{\},codes={\catcode`$=3\catcode`_=8}}
\DefineVerbatimEnvironment{Highlighting}{Verbatim}{commandchars=\\\{\}}

\usepackage{framed}
\definecolor{shadecolor}{RGB}{248,248,248}
\newenvironment{Shaded}{\begin{snugshade}}{\end{snugshade}}

\newcommand{\KeywordTok}[1]{\textcolor[rgb]{0.13,0.29,0.53}{\textbf{{#1}}}}

\newcommand{\CommentTok}[1]{\textcolor[rgb]{0.56,0.35,0.01}{\textit{{#1}}}}

\newcommand{\NormalTok}[1]{{#1}}

\usepackage{microtype}

\aclfinalcopy 

\title{An NLG pipeline for a legal expert system: a work in progress}

         \author{
           {\bf Inari Listenmaa} \\
           {\bf Jason Morris} \\
           {\bf Alfred Ang} \\
           {\bf Maryam Hanafiah} \\
           {\bf Regina Cheong} \\
        \texttt{\small \{ilistenmaa, jmorris, alfredang, maryammh, reginacheong\}@smu.edu.sg} \\
         Singapore Management University}

\date{}

\begin{document}
\maketitle

\begin{abstract}

We present the NLG component for L4, a prototype domain-specific language (DSL) for drafting laws and contracts.
As a concrete use case, we describe a pipeline for a legal expert system created from L4 code.
The NLG component is used in two steps. The first step is to create an interview,
whose answers are processed into a query for an automated reasoner.
The second step is to render the answers of the reasoner in natural language.
\end{abstract}

\section{Introduction}
\label{sec:intro}
We introduce L4, a prototype\footnote{L4 is a work in progress, and this article presents a snapshot of the project as of June 2021. Any concrete examples of L4 code may change in a few months.}
domain-specific language (DSL) for drafting laws and contracts.
L4's applied focus places it within the ``Rules as Code'' movement (e.g. \citet{openfisca_openfisca_nodate}, Catala~\cite{merigoux2021catala}) that itself draws on early computational law thinking \cite{sergot_british_1986, love_computational_2005}. But rather than focusing on encoding laws into existing programming languages, we devise an external DSL designed for legal specification.

Drafting in a high-level, declarative DSL makes it possible to separate the legal layer and the application layer. We encode the rules once, and that encoding is a source for further applications, such as legal expert systems, consistency checking, visualisation and natural language generation.
If the rules change, they need to be changed only once at the source, and the applications can be automatically updated.


In this paper, we describe an NLG pipeline from a set of rules drafted in L4 into natural language.
The generated natural language must adapt to different communicative needs: a piece of code may become, for example, a declaration, a prohibition, a condition, an interview question or an answer to a user query.
Because we require flexibility in output, it is important to have a deep understanding of any source material. Section~\ref{sec:cnl_nlg} describes the first step, of acquiring a base for NLG from user's descriptions of classes and predicates---this step is a general prerequisite to all kinds of NLG tasks, not limited to the one we describe here. Section~\ref{sec:nlg} describes a concrete use case of creating a legal expert system from a L4 encoding.
Finally, Section~\ref{sec:future} outlines future work.


\section{L4 example}
\label{sec:example}


As an introduction to L4, we will encode the rules of Rock-Paper-Scissors (RPS) for two players.

\subsection{L4 basics}

\paragraph{Types and values}
First, we declare data types for the entities needed to represent the rules. A round of RPS is a \texttt{Game} with two \texttt{Player}s, each throwing \texttt{Sign}. Then we introduce the three signs: rock, paper and scissors.

\begin{Shaded}
\begin{Highlighting}[]
\KeywordTok{class}                     \CommentTok{\# types}
\NormalTok{  Player}
\NormalTok{  Game}
\NormalTok{  Sign}

\KeywordTok{decl}                     \CommentTok{\# values}
\NormalTok{  Rock : Sign}
\NormalTok{  Paper : Sign}
\NormalTok{  Scissors : Sign}
\end{Highlighting}
\end{Shaded}

\paragraph{Predicates}

Next, we define four predicates, used later in our rules.
The current syntax of L4 is adopted from functional programming languages. 
For example, the type signature for \texttt{Win : Player~$\rightarrow$~Game  $\rightarrow$ Bool} means \textit{``the function Win takes a Player and a Game, and returns a Boolean''}.

\begin{Shaded}
\begin{Highlighting}[]
\KeywordTok{decl}                 \CommentTok{\# predicates}
\NormalTok{  Participate}
\NormalTok{        : Player → Game → Bool}
\NormalTok{  Throw : Player → Sign → Bool}
\NormalTok{  Win   : Player → Game → Bool}
\NormalTok{  Beat  : Sign → Sign → Bool}
\end{Highlighting}
\end{Shaded}

For the purposes of constructing an interview, it is useful to have information on how to group the different predicates.
L4 offers an alternative syntax for the predicates to be fields in the data types, as follows.

\begin{Shaded}
\begin{Highlighting}[]
\CommentTok{# equivalent to the standalone}
\CommentTok{# versions Participate, Throw, Win}
\KeywordTok{class}
\NormalTok{  Player \{}
\NormalTok{    participate : Game → Bool }
\NormalTok{    throw       : Sign → Bool}
\NormalTok{    win         : Game → Bool}
\NormalTok{  \}}
\end{Highlighting}
\end{Shaded}

The class definition with the fields \texttt{throw}, \texttt{win} and \texttt{participate} is equivalent to the stand-alone predicates in capital letters.
Note that capital letters have no significance in L4, we just use them in this example to distinguish between standalone predicates and those in a field.

As an alternative, \texttt{participate} and \texttt{win} can be defined as fields of the class \texttt{Game}.
With this grouping, the type signature becomes \texttt{Game $\rightarrow$ Player $\rightarrow$ Bool}.
We will use this grouping in the examples in Section~\ref{sec:DAinterview}, because that makes the best order for asking the questions.

\begin{Shaded}
\begin{Highlighting}[]
\KeywordTok{class}        \CommentTok{# different grouping}
\NormalTok{  Game \{}
\NormalTok{    participate : Player → Bool}
\NormalTok{    win         : Player → Bool}
\NormalTok{  \}}
\end{Highlighting}
\end{Shaded}

\paragraph{Rules}
We formulate the rules for winning a game: both players must throw a sign, and the winning player's sign must beat the other player's sign. The syntax for function application follows functional programming conventions: to say that \textit{``x wins rps''}, we write \texttt{Win x rps}.
The relations between the signs are also encoded as rules---implementation omitted for brevity.

\begin{Shaded}
\begin{Highlighting}[]
\KeywordTok{rule }\NormalTok{<winner>}             \CommentTok{\# rules}
\KeywordTok{  for}\NormalTok{  a : Player, g : Game,}
\NormalTok{       r : Sign, s : Sign}
\KeywordTok{  if  exists}\NormalTok{ b : Player }\KeywordTok{.}
\NormalTok{        Participate a g }\KeywordTok{\&\&}
\NormalTok{        Participate b g }\KeywordTok{\&\&}
\NormalTok{        Throw a r }\KeywordTok{\&\&}\NormalTok{ Throw b s }\KeywordTok{\&\&}
\NormalTok{        Beat r s}
\KeywordTok{  then} \NormalTok{Win a g}
\end{Highlighting}
\end{Shaded}

\noindent



\subsection{Natural language descriptions}
\label{sec:NLdesc}

For NLG purposes, the L4 file may include optional descriptions in controlled natural language (CNL).
If no natural language description is given, the predicate name is used instead.
For example, we can record that \texttt{Participate} takes its object with the preposition ``in''.

\begin{Shaded}
\begin{Highlighting}[]
\KeywordTok{lexicon}
\NormalTok{  Participate @ "participate in"}
\end{Highlighting}
\end{Shaded}

The current default is to include only the predicate, and assume the argument order subject--object--indirect object. Subject agreement doesn't matter: ``participate in'' and ``participates in'' are equivalent.
For higher than ternary predicates, alternative argument order or several prepositions, the types may used as placeholder arguments.
This is how to avoid NLG like ``game wins player'', when the predicates are grouped as fields in the classes.

\begin{Shaded}
\begin{Highlighting}[]
\KeywordTok{class}
\NormalTok{  Game \{win : Player → Bool\}}
\CommentTok{  # normalised into predicate}
\CommentTok{  # win : Game → Player → Bool}
\KeywordTok{lexicon}
\NormalTok{  win @ "[Player] wins [Game]}
\end{Highlighting}
\end{Shaded}


\section{CNL to NLG}
\label{sec:cnl_nlg}

\citet{authors_under_submission_2021} describes in detail the CNL and the process of parsing the descriptions, including ambiguity resolution,
so we include only a short version here.

\paragraph{CNL} As seen in Section~\ref{sec:NLdesc}, the L4 classes and predicates can be enriched with a natural language description.
These descriptions are written in a controlled natural language (CNL) which is implemented in Grammatical Framework (GF, \citeauthor{ranta_grammatical_2004} \citeyear{ranta_grammatical_2004}), a programming language for multilingual grammar applications.

Our CNL is based on the GF Resource Grammar Library (RGL) \cite{ranta2009gf}, which provides a library of syntactic structures and morphology for over 30 languages, and a large, multilingual morphological lexicon \cite{angelov-2020-parallel} based on Princeton WordNet \cite{fellbaum1998wordnet},
enriched with syntactic features like valency.

On top of the RGL and the GF-WordNet lexicon, we have added a set of constructions common to legal text.
In addition, we have relaxed the rules for forming sentences, so that we can parse predicates without arguments,
(\textit{participates in}),
or arguments in brackets (\textit{[Player] wins [Game]}).

So far we have only used English, but the method is scalable to any of the languages that are in the RGL and have a large lexicon.




\paragraph{Goal of the CNL} The purpose of our CNL is to be able to parse the user input correctly, and thus use it correctly in the NLG.
Unlike many other CNLs out there
(see \citet{kuhn2014survey} for a survey),
ours is not concerned with well-defined semantics---L4, being a programming language, is much better suited for that. We only care about well understood syntactic structure, so that we can have flexible NLG for different communicative needs.

\section{NLG pipeline for an expert system}
\label{sec:nlg}

We construct an interview in order to find out who wins a game of RPS. The main puzzle, \textit{who wins RPS}, is not one of the questions we pose to the user---instead, we ask about all the prerequisite information, and give the user's answers to an automated reasoner, which will announce the winner.

NLG-wise, this involves two steps. In the first step, we generate interview questions from the L4 code (Section~\ref{sec:DAinterview}).
In the second step, we generate natural language from the output of the reasoner (Section~\ref{sec:SCASPans}).

\subsection{Interview questions}
\label{sec:DAinterview}

The RPS interview will ask about the players and which signs they threw.


{\it
\begin{enumerate}
\item Is there a game?
\begin{itemize}
    \item Answer options: yes/no
    \item Answered: yes
\end{itemize}
\item Who participates in the game?
\begin{itemize}
    \item Answer options:
        \begin{itemize}
          \item slot for free text
          \item loop "Are there more players?"
        \end{itemize}
    \item Answered:
    \begin{itemize}
        \item Alice, yes for more players
        \item Bob, no for more players
    \end{itemize}
\end{itemize}
\item Which sign does Alice throw?
\begin{itemize}
    \item Answer options: rock/paper/scissors
\end{itemize}
\item Which sign does Bob throw?
\begin{itemize}
    \item Answer options: rock/paper/scissors
\end{itemize}
\end{enumerate}
}


\paragraph{Docassemble}

The questions are embedded in a \citeauthor{docassemble} interview. Docassemble is an open-source legal expert system, where users answer a set of questions through a browser interface.
The responses are then compiled together into a document 
or processed further in other applications---in our case, an automated reasoner.

In order to let the user customise the interview---for instance, the NLG, question ordering, additional links to source material---
our system outputs a LExSIS ({\bf L}egal {\bf Ex}pert {\bf S}ystem {\bf I}nterface {\bf S}chema) file, which is short and declarative.
The final interview, a long and imperative piece of Python code, is then constructed from the LExSIS file.


\paragraph{Types of questions}

The interview includes three types of questions: a yes--no question (1), an open-ended wh-question (2) and two enumeration questions (3,4).
These are determined by the L4 code. 

\texttt{Game} is a class, so the first question we ask is whether a game exists.
When we know that a game exist, we ask wh-questions about the predicates related to the game: \textit{who participates} and \textit{which sign} do they throw.
The L4 code lists three signs, but no pre-existing players: that's why the answer options for (2) is a slot for free text, and for (3,4) an enumerated list.

The choice of \textit{who} in (2) is because \textit{player} is a human noun. The GF-WordNet lexicon \cite{angelov-2020-parallel} contains this information for about 6500 words, so we just make it as a rule in the grammar. If some human nouns are missing the animacy information, they will just be verbalised with the general strategy, \textit{which players}. 


\paragraph{Order of questions}

The order of the questions is determined by Docassemble and the L4 encoding.

Docassemble has its internal logic: the goal of the interview is this case to find out the winner of a game of RPS, and thus it starts out from the first prerequisite, \textit{is there a game}?

L4 too has its own logic: if we used the encoding where \texttt{participate} is a field of the class \texttt{Game}, then it will group the questions accordingly. Participation is a feature of a game, so after ensuring that a game exists, the user is asked about the game's participants---\textit{game} is now old information, so in English it gets a definite article and is placed later in the question.
With a different grouping of predicates in L4, the questions will also be grouped and phrased differently: for instance, \textit{Are there any players}, followed by \textit{Which game does Alice participate in}.





\paragraph{Constructing the questions}
Table~\ref{tab:gfExps} shows the GF expressions needed to construct different types of sentences.
All of the functions in normal typeface come from the language-independent API of the GF RGL\footnote{ {\tiny \url{http://www.grammaticalframework.org/lib/doc/synopsis/}}}.
The variables in italics are generic: any noun phrase could be in place of {\tt \textit{np}}, any transitive verb for {\tt \textit{v2}}. For a verb like \textit{participate in}, the preposition is part of the verb's lexical entry---that's why the expression to construct a VP is identical for \textit{participates in RPS} and \textit{throws rock}.

None of the expression templates which includes a noun phrase specifies a determiner.
The NP can be constructed in different phases: it can be new or old information, or it may be a proper noun. Unfortunately, the lexicon doesn't contain properties like mass noun vs. count noun, nor idiomatic expressions, like \textit{throw rock} instead of \textit{throw a rock}. Currently, we just accept that sometimes the determiner choice is unnatural, and leave it to the user to postprocess the output.






\subsection{Reasoner output in natural language}
\label{sec:SCASPans}

The second output target is a verbalisation of the output from an automated reasoner.
Using the user answers to the interview from Section~\ref{sec:DAinterview}, 
a solution satisfying all the constraints is generated.
The solution is then restructured in natural language, becoming the conclusion the user receives.\\


{\it
Alice wins RPS, because
\begin{itemize}
\item Alice throws paper and Bob throws rock, and
\item paper beats rock.
\end{itemize}
}

\paragraph{\scasp query}

The reasoner we use is \scasp \cite{arias2018constraint}: a logic programming language for Constraint Answer Set Programming.

The L4 code is translated into \scasp rules and statements.
The query ``who wins a game of RPS'' is phrased as {\tt ?-~win(Game,Player)}, where Player and Game are \scasp variables.
The order of the arguments comes from the L4 encoding where {\tt win} is a field of the class {\tt Game}.

The L4 encoding doesn't list any players or games,
so \scasp needs more information before it can meaningfully run the query. As we have seen in \ref{sec:DAinterview}, this information comes from the interview.



\paragraph{\scasp answers}
The answer consists of one or more sets of \scasp statements.
These statements, just like their source in L4, are connected to the CNL descriptions, which are stored as GF abstract syntax trees. This makes it possible to aggregate them, such as grouping them by subject or predicate.
In case of multiple sets for the same query, we group repeating statements in one block, grouped by {\it and}, and the rest under another, grouped by {\it or}.
An example of this strategy is shown in Table~\ref{tab:aggregation}, which shows an answer to a slightly different query: ``list all ways that Alice can win a game''.

Currently, our order of aggregating predicates is intransitive before transitive, and nouns/adjectives before verbs. This results in sentences like
{\it are players and participants in RPS},
{ \it are players and participate in RPS},
{ \it play and participate in RPS},
or less fluently, {\it are participants in RPS and play}.

\section{Future work}
\label{sec:future}

As of June 2021, the whole L4 system, including the NLG component, is just a prototype.
Future work for the CNL and user descriptions is listed in \citet{authors_under_submission_2021}.
The quality of NLG depends on the quality of user input---no amount of sophisticated information structure will cover for badly parsed description, so improving the CNL is our first priority.

We have demonstrated in this paper the system's capability to answer questions of type {\it What Is True} (Alice wins RPS), {\it When Is This True} (when Alice throws paper and Bob throws rock) and {\it Why} (because paper beats rock). As we scale up to larger examples, we will need more fine-grained heuristics in aggregation, information structure and ordering, as well as adding variation to the structures. We also plan to add more languages. 


In the longer term, we plan to support more types of queries, with hypotheticals and goal-seeking, such as {\it What Would Make This True} and {\it What Else Is True}, and new types of questions and answers require new NLG functions.
As another long-term goal, we would like to incorporate more semantic knowledge in our NLG.





\begin{table*}[t]
\centering
\caption{GF expressions to construct different questions}
\begin{tabular}{|p{0.27\textwidth}|p{0.685\textwidth}|}
\hline
Is there a game & { \tt mkQS (mkCl \textit{noun}) }
  \\
\hline
Does Bob participate in RPS & {\tt mkQS (mkCl \textit{np} (mkVP \textit{v2 np})) }
  \\
\hline
Who participates in RPS & {\tt mkQS (mkQCl who\_IP (mkVP \textit{v2 np})) }
  \\
\hline
Which player throws rock & {\tt mkQS (mkQCl (mkIP which\_IDet \textit{noun}) (mkVP \textit{v2 np})) }
  \\
\hline
Which sign does Alice throw  & {\tt mkQS (mkQCl (mkIP which\_IDet \textit{noun}) (mkClSlash \textit{np} (mkVPSlash \textit{v2}))) }
 \\
\hline
\end{tabular}
\label{tab:gfExps}
\\
\caption{Aggregation of multiple models: answer to all ways Alice can win}
\begin{tabular}{|p{0.02\textwidth}p{0.935\textwidth}|}
\hline
 \multicolumn{2}{|l|}{Alice wins RPS, { if all of the following hold:}}  \\ \hline
   & RPS is a game, and  \\
   & Alice and Bob are players and participate in RPS \\ \hline
\multicolumn{2}{|l|}{ and one of the following holds:}  \\ \hline
 & rock beats scissors, Alice throws rock and Bob throws scissors, \\ \hline
 & scissors beats paper, Alice throws scissors and Bob throws paper, or \\ \hline
 & paper beats rock, Alice throws paper and Bob throws rock.   \\ \hline
\end{tabular}
\label{tab:aggregation}
%
\end{table*}

\bibliographystyle{acl_natbib}
\bibliography{main}

\end{document}